\documentclass[final]{cvpr}

\usepackage{times}
\usepackage{epsfig}
\usepackage{graphicx}
\usepackage{amsmath}
\usepackage{amssymb}
\usepackage{subfig}
\usepackage{algorithm}
\usepackage{algorithmic}
\newcommand{\sign}{\text{sign}}
\newcommand{\tabincell}[2]{\begin{tabular}{@{}#1@{}}#2\end{tabular}}

\usepackage[pagebackref=true,breaklinks=true,colorlinks,bookmarks=false]{hyperref}

\def\cvprPaperID{10} 
\def\confYear{CVPR 2021}

\begin{document}
	
	\title{Fast Walsh-Hadamard Transform and Smooth-Thresholding Based Binary Layers in Deep Neural Networks}
	
	\author{Hongyi Pan\\
		\and
		Diaa Badawi\\
		\and
		Ahmet Enis Cetin\\
		\and
		Department of Electrical and Computer Engineering\\
		University of Illinois at Chicago\\
		{\tt\small \{hpan21, dbadaw2, aecyy\}@uic.edu}\\
	}
	\maketitle

	\begin{abstract}
		In this paper, we propose a novel layer based on fast Walsh-Hadamard transform (WHT) and smooth-thresholding to replace $1\times 1$ convolution layers in deep neural networks. In the WHT domain, we denoise the transform domain coefficients using the new smooth-thresholding non-linearity, a smoothed version of the well-known soft-thresholding operator. We also introduce a family of multiplication-free operators from the basic 2$\times$2 Hadamard transform to implement $3\times 3$ depthwise separable convolution layers. Using these two types of layers, we replace the bottleneck layers in MobileNet-V2 to reduce the network's number of parameters with a slight loss in accuracy. For example, by replacing the final third bottleneck layers, we reduce the number of parameters from 2.270M to 947K. This reduces the accuracy from 95.21\% to 92.88\% on the CIFAR-10 dataset. Our approach significantly improves the speed of data processing. The fast Walsh-Hadamard transform has a computational complexity of $O(m\log_2 m)$. As a result, it is computationally more efficient than the $1\times1$ convolution layer. The fast Walsh-Hadamard layer processes a tensor in $\mathbb{R}^{10\times32\times32\times1024}$  about 2 times faster than $1\times1$ convolution layer on NVIDIA Jetson Nano computer board.
	\end{abstract}
	
	\section{Introduction}
	
	Deep convolution neural networks (CNN) are universally used in a wide range of applications including image classification~\cite{krizhevsky2012imagenet,simonyan2014very, szegedy2015going, wang2017residual, he2016deep, badawi2020computationally, agarwal2021coronet, partaourides2020self, stamoulis2018designing}, object detection~\cite{redmon2016you, aslan2020deep, menchetti2019pain, aslan2019early} and semantic segmentation~\cite{yu2018bisenet, huang2019ccnet, long2015fully, poudel2019fast, jin2019fast}. On the other hand, implementing deep neural networks in real-time resource-constrained environments such as embedded devices is very difficult due to insufficient memory and limited computational capacity. 
	For these reasons, there is an increasing demand to develop smaller and efficient neural networks as neural networks have become an instrumental technology for sensor data processing \cite{schwenk2000boosting} and can be utilized in our modern society everywhere.
	
	Efficient neural network models include compressing a large neural network using quantization~\cite{wu2016quantized,muneeb2020robust}, hashing~\cite{chen2015compressing}, pruning~\cite{pan2020computationally}, vector quantization~\cite{yu2018gradiveq} and Huffman encoding~\cite{han2019deep}. Another approach is the  SqueezeNet~\cite{iandola2016squeezenet}, which is designed as a small network with $1\times 1$ filters during training.  In~\cite{ayi2020rmnv2} the neural network is slimmed by removing some layers from a well-developed model.
	Binary neural networks~\cite{courbariaux2016binarized, bulat2018hierarchical, rastegari2016xnor, shen2021s2, liu2020reactnet, martinez2020training, bulat2020bats, hubara2016binarized, alizadeh2018empirical, bannink2020larq} showed that binary weights can be used to slim and accelerate neural networks. 
	
	Although  $1\times1$ convolutions reduce the computational load, they are still computationally expensive and time-consuming in regular deep neural networks.
	In this paper, we introduce a binary layer based on the fast Walsh-Hadamard transform to slim and speed up deep neural networks with $1\times1$ convolutions. Moreover, the recent literature \cite{akbacs2015multiplication, afrasiyabi2017energy, afrasiyabi2018non, badawi2017multiplication, pan2019additive} developed an energy-efficient neuron called multiplication-free (MF) kernel, which does not require any multiplications. We establish the relation between the MF operator and 2-by-2 Hadamard transform, and we fuse this idea to propose the depthwise separable multiplication-free convolution layer. We apply these two types of layers in bottleneck layers of MobileNet-V2~\cite{sandler2018mobilenetv2}, and the new network is remarkably more slimmed and computationally efficient compared to the original structure according to our experiments. Similar results can also be obtained in other standard deep neural networks.
	
	\section{Methodology}\label{Methodology}
	In this section, we first review the Walsh-Hadamard (WH) transform. Then, we introduce the proposed fast Walsh-Hadamard transform layer.  Finally, we describe the multiplication-free depthwise separable convolution layer, which turns out to be a combination of two binary layers. 
	\subsection{The Walsh-Hadamard (WH) Transform}
	The WH transform is based on the so-called butterfly operation described by the following binary matrix:
	\begin{equation}
		\mathbf{H} =
		\begin{bmatrix}
			1&1\\1&-1\\
		\end{bmatrix}
	\end{equation}
	which is also the main building block of the fast Fourier transform (FFT). A 2-by-2 WH transform $\bf{Y}$ of
	the vector $\mathbf{X}\in \mathbb{R}^2$ is
	\begin{equation}
		\mathbf{Y} = \mathbf{W}\mathbf{X} = \mathbf{H} \mathbf{X} 
	\end{equation}
	In general, the WH transform $\mathbf{Y} = \mathbf{W}_k\mathbf{X}$ of a vector $\mathbf{X} \in \mathbb{R}^m$ where $m=2^k, k \in \mathbb{N}$ can be expressed via the orthogonal Walsh matrix $\mathbf{W}_k\in \mathbb{R}^{m\times m}$ which is generated using the Hadamard matrix which can be recursively constructed in two steps as follows \cite{walsh1923closed}:
	\begin{itemize}
		\item First, we construct the Hadamard matrix  $\mathbf{H}_k$:
		\begin{equation}
			\mathbf{H}_k = 
			\begin{cases}
				1,& k = 0,\\
				\begin{bmatrix}
					\mathbf{H}_{k-1} & \mathbf{H}_{k-1} \\ \mathbf{H}_{k-1} & -\mathbf{H}_{k-1}
				\end{bmatrix},& k > 0,
			\end{cases}
		\end{equation}
		Alternatively, for $k>1$, $\mathbf{H}_k$ can also be computed as 
		\begin{equation}
			\mathbf{H}_k=\mathbf{H}_1 \otimes\mathbf{H}_{k-1}
		\end{equation}
		where $\otimes$ denotes Kronecker product.
		\item Then, we shuffle the rows of $\mathbf{H}_k$ to obtain $\mathbf{W}_k$ by applying the bit-reversal permutation and the Gray-code permutation on row index. For example, when $k=2, m=4$, 
		we have the Hadamard matrix
		\begin{equation}
			\mathbf{H}_2 =
			\begin{bmatrix}
				1&1&1&1\\1&-1&1&-1\\1&1&-1&-1\\1&-1&-1&1\\
			\end{bmatrix}
		\end{equation}
		and the corresponding WH-transform is given by
		\begin{equation}
			\mathbf{W}_2 =
			\begin{bmatrix}
				1&1&1&1\\1&1&-1&-1\\1&-1&-1&1\\1&-1&1&-1\\
			\end{bmatrix}
		\end{equation}
	\end{itemize}
	It is not difficult to verify that  $\mathbf{X} = \frac{1}{m}\mathbf{W}_k\mathbf{Y}$, which implies the inverse Walsh-Hadamard transform is itself with normalization by $m$.
	
	The fast Walsh-Hadamard transform (FWHT) algorithm is similar to the FFT algorithm, and, consequently, the complexity of FWHT is $O(m\log_2m)$. The FWHT algorithm is completely based on the butterfly operations described in Eq. (1) in~\cite{fino1976unified}. As a result, the implementation of FWHT can be realized by only addition and subtraction operations using butterflies. 
	It was shown that the WH transform is the same as the block Haar wavelet transform \cite{cetin1993block}.
	
	\subsection{The FWHT Layer}
	\label{FWHT Layer}
	In the state-of-art deep convolution neural networks, $1\times 1$ convolution layers are widely used as hidden layers~\cite{he2016deep, sandler2018mobilenetv2, xie2017aggregated, szegedy2017inception} to change the dimensions of channels. For instance, there are 17 bottleneck layers in MobileNet-V2 \cite{sandler2018mobilenetv2}. The first bottleneck layer contains a $3\times3$ depthwise separable convolution layer and a $1\times1$ convolution layer. Other bottleneck layers are built using a $1\times1$ convolution layer named ``conv\_expand", a $3\times3$ depthwise separable convolution layer~\cite{chollet2017xception}, and a $1\times1$ convolution layer named ``conv\_project". The layer ``conv\_expand" increases the number of channels 6 times. On the other hand, the layer ``conv\_project" decreases the number of channels by a factor of $\frac{1}{6}$. The depthwise separable convolution layer is the main component for feature extraction.
	
	Notice that each $1\times1$ convolution layer contains as many parameters as the number of channels, and they are time-consuming during inference. We propose a novel FWHT layer to replace the $1\times1$  convolution layers. The FWHT layer is summarized in Algorithms~\ref{al: FWHT layer expand} and~\ref{al: FWHT layer project}. In general, an FWHT layer consists of an FWHT to change the tensor to the Hadamard domain, a smooth-thresholding operation as the non-linearity in the Hadamard domain, and an FWHT to change the tensor back to the feature-map domain. The FWHT is performed on the channel axis, which implies that completing FWHT on a tensor $\mathbf{X}\in\mathbb{R}^{n\times w\times h\times m}$ means performing $n\times w\times h$ $m$-length FWHTs in parallel. 
	
	In the WH transform domain, we denoise the parameters and eliminate coefficients with small amplitude values.
	Soft-thresholding (ST) is commonly used in wavelet domain denoising algorithms \cite{agante1999ecg, donoho1995noising} and defined as
	\begin{equation}
		y = \text{S}_T (x) = \sign(x)(|x|-T)_+ = \begin{cases}
			x+T, & x < -T\\
			0, & |x| \le T\\
			x-T, & x > T
		\end{cases}
	\end{equation}
	where $T$ is the thresholding parameter and is trainable in our FWHT layer. 
	
	The denoising parameter $T$ can be learned using the back-propagation algorithm during training. More specifically, 
	\begin{equation}
		\frac{\partial(\sign(x)(|x|-T)_+)}{\partial T} = \begin{cases}
			1, & x < -T\\
			0, & |x| \le T\\
			-1, & x > T
		\end{cases}
		\label{eq: dstdT}
	\end{equation}
	However, the denoising parameter $T$ in soft-thresholding can only be updated by $+1$ or $-1$.  We introduce smooth-thresholding by revising the definition of the soft-thresholding operator as follows
	\begin{equation}
		y = \text{S}_T' (x) = \tanh(x)(|x|-T)_+
	\end{equation}
	and the corresponding partial derivative is
	\begin{equation}
		\frac{\partial(\tanh(x)(|x|-T)_+)}{\partial T} = \begin{cases}
			-\tanh(x), & |x| > T\\
			0, & |x| \le T\\
		\end{cases}
		\label{eq: dttdT}
	\end{equation}
	We observe that the derivative in Eq.~(\ref{eq: dttdT}) is the derivative in Eq.~(\ref{eq: dstdT}) multiplied by $\tanh(x)$. As a result, the convergence of the smooth-thresholding operator is smooth and steady in the back-propagation algorithm. We learn a different threshold $T$ value for each WH transform domain coefficient. 
	Other methods using the Hadamard transform include~\cite{deveci2018energy, zhao2019building} but they did not apply trainable thresholding in the WH-transform domain.
	
	\begin{figure}[htbp]
		\centering
		\includegraphics[width=1\linewidth]{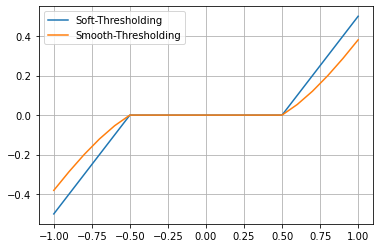}
		\caption{$y=\text{S}_T(x, 0.5)$ and $y=\text{S}'_T(x, 0.5)$}
		\label{fig: st}
	\end{figure}
	Figure~\ref{fig: st} shows the soft-thresholding and smooth-thresholding operators for $T=0.5$, respectively.
	The FWH transform coefficients can take both positive and negative values. Large positive and negative transform domain coefficients are equally important.  Therefore, we cannot use the ReLU function in the transform domain. In fact, our experiments verify this observation, and soft-thresholding and smooth-thresholding improve the recognition accuracy compared to the ReLU.
	
	\begin{algorithm}[htbp]
		\caption{The FWHT layer for channel expansion}
		\begin{algorithmic}[1]
			\renewcommand{\algorithmicrequire}{\textbf{Input:}}
			\renewcommand{\algorithmicensure}{\textbf{Output:}}
			\REQUIRE Input tensor $\mathbf{X}\in\mathbb{R}^{n\times w\times h\times c}$
			\ENSURE  Output tensor $\mathbf{Z}\in\mathbb{R}^{n\times w\times h\times tc}$
			\STATE Find minimum $d\in \mathbb{N}$, s.t. $2^d\ge tc$ 
			\STATE $\hat{\mathbf{X}} = \text{pad}(\mathbf{X}, 2^d-c) \in\mathbb{R}^{n\times w\times h\times 2^d}$
			\STATE $\mathbf{Y} = \text{FWHT}(\hat{\mathbf{X}}) \in\mathbb{R}^{n\times w\times h\times 2^d}$
			\STATE $\hat{\mathbf{Y}} = \text{concat}(\mathbf{Y}[:, :, :, 0], \text{ST}(\mathbf{Y}[:, :, :, 1:]))$
			\STATE $\hat{\mathbf{Z}} = \text{FWHT}(\hat{\mathbf{Y}}) \in\mathbb{R}^{n\times w\times h\times 2^d}$
			\STATE $\mathbf{Z} = \hat{\mathbf{Z}}[:, :, :tc]$
			\RETURN $\mathbf{Z}$.\\
			Comments: Function pad($\mathbf{A}, b$) pads $b$ zeros on the channel axis of tensor $\mathbf{A}$. FWHT($\cdot$) is the normalized fast Walsh-Hadamard transform on the last axis. Function concat($\cdot, \cdot$)  concatenates two tensors along the last axis. ST($\cdot$) performs smooth-thresholding. Index follows Python's rule.
		\end{algorithmic}
		\label{al: FWHT layer expand}
	\end{algorithm}
	
	\begin{algorithm}[htbp]
		\caption{The FWHT layer for channel projection}
		\begin{algorithmic}[1]
			\renewcommand{\algorithmicrequire}{\textbf{Input:}}
			\renewcommand{\algorithmicensure}{\textbf{Output:}}
			\REQUIRE Input tensor $\mathbf{X}\in\mathbb{R}^{n\times w\times h\times tc}$
			\ENSURE  Output tensor $\mathbf{Z}\in\mathbb{R}^{n\times w\times h\times c}$
			\STATE Find minimum $p, q\in \mathbb{N}$, s.t. $2^p\ge tc, 2^q \ge c$ 
			\STATE $r = 2^{p-q}$
			\STATE $\hat{\mathbf{X}} = \text{pad}(\mathbf{X}, 2^p-tc) \in\mathbb{R}^{n\times w\times h\times 2^p}$
			\STATE $\mathbf{Y} = \text{FWHT}(\hat{\mathbf{X}}) \in\mathbb{R}^{n\times w\times h\times 2^p}$
			\STATE $\hat{\mathbf{Y}} = \text{concat}(\mathbf{Y}[:, :, :, 0]/r, \text{avgpool}(\text{ST}(\mathbf{Y}[:, :, :, 1:2^p-r+1]), r))\in\mathbb{R}^{n\times w\times h\times 2^q}$
			\STATE $\hat{\mathbf{Z}} = \text{FWHT}(\hat{\mathbf{Y}}) \in\mathbb{R}^{n\times w\times h\times 2^q}$
			\STATE $\mathbf{Z} = \hat{\mathbf{Z}}[:, :, :c]$
			\RETURN $\mathbf{Z}$.\\
			Comments: Function pad($\mathbf{A}, b$) pads $b$ zeros on the channel axis of tensor $\mathbf{A}$. FWHT($\cdot$) is the normalized fast Walsh-Hadamard transform on the last axis. Function concat($\cdot, \cdot$) concatenates two tensors along the last axis. Function avgpool($\mathbf{A}, b$) is the average pooling on $\mathbf{A}$ with pooling size and strides are $b$. ST($\cdot$) performs smooth-thresholding. Index follows Python's rule.
		\end{algorithmic}
		\label{al: FWHT layer project}
	\end{algorithm}
	
	There are $(2^d-1)$ trainable thresholding parameters in the FWHT layer when there are $2^d$ channels. This means that if we perform smooth-thresholding on tensor $\mathbf{Y}$, each slice among the channel axis will share a common threshold value. The so-called DC channel $\mathbf{Y}[:, :, :, 0]$ usually contains essential information. Due to this reason, we do not perform smooth-thresholding on this channel. 
	
	The FWHT layer for channel expansion is described in Algorithm~\ref{al: FWHT layer expand}.  We first compute the $2^d$ point WH transforms and perform smooth-thresholding in the transform domain. We pad $(2^d-c)$ zeros to the end of each input vector before the $2^d$-by-$2^d$ WH transform to increase the dimension. After smooth-thresholding in the WH domain, we calculate the inverse WH transform. 
	
	In FWHT layer for channel projection by a factor of $r= \frac{2^p}{2^q}$, we first compute the $2^p$ point WH transforms and perform smooth-thresholding in the transform domain as described in Algorithm \ref{al: FWHT layer project}. After this step, we compute the $2^q$ point WH transforms to reduce the dimension of the feature map. We divide the DC channel values by $r$ to keep the energy at the same level as other channels after pooling. In Step 5 of Algorithm \ref{al: FWHT layer project}, we average pool the transform domain coefficients to reduce the dimension of the Walsh-Hadamard transform and discard the last $(r-1)$ transform domain coefficients of $\mathbf{Y}$ to make the dimension equal to $2^q$. The last $(r-1)$ coefficients are high-frequency coefficients, and usually, their amplitudes are negligible compared to other WH coefficients.
	
	Therefore, the dimension change operation from $m$ dimensions to $n$ dimensions can be summarized as follows:
	\begin{equation}
		\mathbf{Z} = \begin{cases}
			\frac{1}{2^{q}}\mathbf{U}\mathbf{W}_q\mathbf{S'_T}\mathbf{W}_q\mathbf{P}\mathbf{X},  \ \ \ m \le n\\
			\frac{1}{\sqrt{2^{p+q}}}\mathbf{U}\mathbf{W}_q\mathbf{A_{vg}}\mathbf{S'_T}\mathbf{W}_p\mathbf{P}\mathbf{X}, \ \  m > n
			\label{eq: FWHT layer}
		\end{cases}
	\end{equation}
	where $p$ is the minimum integer such that $2^p\ge m$, $q$ is the minimum integer such that $2^q\ge n$, $\mathbf{P}$ describes zero-padding operation to make $\mathbf{X}$ multipliable by $\mathbf{W}_p$ or $\mathbf{W}_q$, $\mathbf{S'_T}$ is the smooth-thresholding layer with DC channel excluded, $\mathbf{U}$ is unpadding function to make the dimension the same as $\mathbf{Z}$, and $\mathbf{Avg}$ is average pooling on the channel axis with the DC channel of WH transform excluded. 
	
	Hence, the trainable number of parameters in FWHT layers is no more than the $(2^d-1)$ where $d$ is the minimum integer such that $2^d$ is no less than the number of input channels. The trainable parameters are only the threshold values of the smooth-thresholding. Therefore, it is clear that FWHT layer requires significantly fewer parameters than the regular $1\times1$  layer, which requires a different set of filter coefficients for each $1\times 1$ convolution.
	
	\subsection{Multiplication-Free Depthwise Separable "Convolutions" (MF-DS-Conv)}
	Akba{\c{s}} {\em et.\ al.} introduced multiplication-free (MF) kernel to replace regular convolution in CNNs \cite{akbacs2015multiplication, afrasiyabi2017energy, afrasiyabi2018non, badawi2017multiplication, pan2019additive}. The MF kernel requires no multiplication but only additions and sign operations. It is more energy-efficient compared to the standard convolution.  In this section, we describe how we can generate new operators similar to the MF kernel from the 2$\times$2 Walsh-Hadamard transform to implement depthwise separable convolution using binary operations, additions, and subtractions.
	
	Let $w$ be a weight value and $x$ be the corresponding input. In a regular neuron we multiply $x$ by $w$ to determine the contribution of the input value. In some binary networks and additive networks \cite{akbacs2015multiplication, afrasiyabi2017energy, afrasiyabi2018non, badawi2017multiplication, pan2019additive}, the sign of the multiplication $w \times x$ is used as the basic operation of the neuron. In addition, we will scale the $\sign(w \times x)$ by the  $\ell_1$ norm of the vector $y=[w+x \ \ w- x]^T$ which is the 2$\times$2 WH transform of the vector 
	$[w \ \ x]^T$ and define the following basic operation
	\begin{equation}
		\begin{array}{ll}
			w \odot x  & = \sign(w\times x)(|w+x| + |w-x|)  \\
			& = 2\sign(w \times x) \max\{|w|, |x|\} 
		\end{array}
		\label{min}
	\end{equation}
	which can be used to construct a vector "product". The operator can also replace the multiplication operator used in correlation or convolution calculations. Another related operator is
	\begin{equation}
		\sign(w \times x )(||w+x| - |w-x||) = 2\sign(w \times x)\min\{|w|, |x|\}
		\label{max}
	\end{equation}
	which can be also used in convolution and correlation operations.
	The above operations defined in Eq. (\ref{min}) and (\ref{max}) are very similar to the MF operation used in 
	\cite{akbacs2015multiplication, afrasiyabi2017energy, afrasiyabi2018non, badawi2017multiplication, pan2019additive}, which is 
	\begin{equation}
		w \oplus x =sign(w\times x)(|w| + |x|) 
		\label{mf}
	\end{equation}
	What is common in all of the above operations is that the result of the operation  $w \oplus x$ has the same sign as the multiplication operation, and the magnitude of the output is determined either by addition, min, or max operations.
	As a result, we can use all of the above three operations defined in Eq. (\ref{min})- (\ref{mf}) in vector product, correlation, and convolution operations as follows:
	Let $\mathbf{w} = [w_1 \cdots w_D]^T \in \mathbb{R}^{D\times 1}$ and $\mathbf{x} = [x_1 \cdots x_D]^T \in \mathbb{R}^{D\times 1}$  be two $D$-dimensional column vectors.  Instead of performing the standard Euclidean inner product 
	$
	\langle \mathbf{w}, \mathbf{x} \rangle = \mathbf{w}^T \mathbf{x} \triangleq \sum_{i=1}^D w_i x_i
	$,
	the MF dot product that is defined as 
	\begin{align}
		\mathbf{w}^T\oplus \mathbf{x} &\triangleq \sum_{i=1}^D \sign(w_i x_i)(|w_i| + |x_i|)\\ &=\sum_{i=1}^D\sign(w_i)x_i+w_i\sign(x_i)
		\label{eqn_mf_vector}
	\end{align}
	is used in neurons and convolutions. Vector ''products'' using Eq. (\ref{min}) and (\ref{max}) can be also defined in a similar manner.
	
	Notice that the only multiplication operations that appear in Eq. (\ref{eqn_mf_vector}) correspond to sign changes and can be implemented with very low complexity. For this reason, we do not count the sign changes towards multiplication operations and thus call Eq. (\ref{eqn_mf_vector}) an MF dot product. It can easily be verified that the product in Eq. (\ref{eqn_mf_vector}) and operators (\ref{min}) and (\ref{max}) induce a scaled version of $\ell_1$-norm as 
	\begin{equation}
		\mathbf{x}^T \oplus \mathbf{x} = \sum_{i=1}^n |x_i|+|x_i| = 2\|\mathbf{x}\|_1
	\end{equation}
	
	Vector dot products described above can be extended to matrix multiplications as follows: Let $\mathbf{W} \in \mathbb{R}^{n\times m}$ and $\mathbf{X} \in \mathbb{R}^{n\times p}$ be arbitrary matrices. Then we define
	\begin{align}
		\mathbf{W^T} \!\oplus\! \mathbf{X} \triangleq 
		\begin{bmatrix}
			\mathbf{w}_1^T\oplus \mathbf{x}_1&\mathbf{w}_1^T\oplus \mathbf{x}_2&\dots&\mathbf{w}_1^T\oplus \mathbf{x}_p\!\!\!\!\!\!\\
			\mathbf{w}_2^T\oplus \mathbf{x}_1&\mathbf{w}_2^T\oplus \mathbf{x}_2&\dots&\mathbf{w}_2^T\oplus \mathbf{x}_p\!\!\!\!\!\!\\
			\vdots&\vdots&\ddots&\vdots&\\
			\mathbf{w}_m^T\oplus \mathbf{x}_1&\mathbf{w}_m^T\oplus \mathbf{x}_2&\dots&\mathbf{w}_m^T\oplus \mathbf{x}_p\!\!\!\!\!\!
		\end{bmatrix}
		\label{eq: matrix}
	\end{align}
	where $\mathbf{w}_i$ is the $i$th column of $\mathbf{W}$ for $ i = 1,\ 2,\ \dots,\ m$ and $\mathbf{x}_j$ is the $j$th column of $\mathbf{X}$ for $j = 1,\ 2,\ \dots,\ p$.  In brief, the definition is similar to the matrix production $\mathbf{W}^T\mathbf{X}$ by only changing the element-wise product to element-wise MF-product. By replacing product operation by MF product operation in depthwise separable convolution layer, we achieve MF-DS-Conv.
	
	The standard back-propagation algorithm can be used for training the MF-DS-Conv layer with the need for small approximations. The partial scalar derivatives of the pre-activation response concerning are given as follows:
	\begin{equation}
		\frac{\partial( w \oplus x)}{\partial x} = \sign(w) + 2w \delta(x)
		\label{formular: dx}
	\end{equation}
	\begin{equation}
		\frac{\partial(w \oplus x)}{\partial w} = 2x \delta(w) + \sign(x)
		\label{formular: dw}
	\end{equation}
	where $\delta(x)$ is the Dirac--delta function that directly results from the discontinuity of the signum function at $x=0$. If we omit the delta function from the definitions of the partial derivatives, we end up with binary derivatives ($\sign(w)$ and $\sign(x)$). However, approximating the Dirac--delta function provides better convergence since we end up with smoother derivatives. In this regard, we approximate the derivative of the signum function to be that of a steep hyperbolic tangent, as follows:
	\begin{equation}
		\frac{d \sign(x)}{d x} \approx \frac{ d \text{tanh}(\alpha x)}{d x} = \alpha \big(1-\text{tanh}^2(\alpha x)\big)
	\end{equation}
	for a scalar $\alpha >> 1$. This is reasonable since $\sign(x) = \lim_{\alpha \to \infty} \text{tanh}(\alpha x)$. This way the terms associated with the delta function in the Eq. (\ref{formular: dx}) and (\ref{formular: dw}) will contribute to the partial derivatives when the arguments are close to zero. In our experiments, we choose $\alpha=10$.
	
	By replacing two $1\times 1$ convolution layers by FWHT layers and the depthwise separable $3\times 3$ convolution layer by a $3\times 3$ MF-DS-Conv layer, we implement ``multiplication-free" bottleneck layer. Although it is not wholly multiplication-free because there are batch-normalization layers between the convolution layers, it is much more efficient and slimmer than the original bottleneck layer. On the other hand, as it is shown in Table~\ref{tab: bottleneck} because the input and the output of each layer have the same shape as the original layer, our layers are plug-and-play.
	
	\begin{table}[htbp]
		\begin{center}
			\begin{tabular}{|c|c|c|}
				\hline
				Input & Operator & Output\\
				\hline\hline
				$h\times w \times k$&FWHT expension, ReLU6& $h\times w \times tk$\\
				$h\times w \times tk$&$3\times3$ MF-DS-Conv, ReLU6& $\frac{h}{s}\times \frac{w}{s} \times tk$\\
				$\frac{h}{s}\times \frac{w}{s} \times tk$&FWHT projection, ReLU6& $\frac{h}{s}\times \frac{w}{s} \times k'$\\
				\hline
			\end{tabular}
		\end{center}	
		\caption{Revised bottleneck residual block transforming from $k$
			to $k'$ channels, with stride $s$, and expansion factor $t$. Batch normalization is used after each layer. Input and output are same as Table 1 in~\cite{sandler2018mobilenetv2}.}
		\label{tab: bottleneck}
	\end{table}
	
	\subsection{Weighted Smooth-Thresholding}
	We further propose the improved version of the smooth-thresholding. \begin{equation}
		y = \text{S}_{WT}(x) = \tanh(wx)(|wx|-T)_+,
		\label{eq: Weighted Smooth-Thresholding}
	\end{equation}
	that is, we multiply the input with a trainable weight. We can extend Eq. (\ref{eq: Weighted Smooth-Thresholding}) in to the matrix form. Let $\mathbf{X}\in\mathbb{R}^{n\times w\times h\times m}$ be the data in the Walsh-Hadamard domain with the Hadamard size $m$, we will have trainable $\mathbf{w}\in\mathbb{R}^m$ and $\mathbf{T}\in\mathbb{R}^m$, then the thresholding can be represent as
	\begin{equation}
		\mathbf{Y} = \text{S}_{W\circ T}(\mathbf{W\circ X}) = \tanh(\mathbf{W\circ X})(|\mathbf{W\circ X}|-\mathbf{T})_+,
	\end{equation}
	,where $\mathbf{W\circ X}$ are the element-wise product between $\mathbf{W}$ and $\mathbf{X}$. We don't apply regular matrix-vector product for computational efficiency and saving parameters. Then, Eq.(\ref{eq: FWHT layer}) becomes 
	\begin{equation}
		\mathbf{Z} = \begin{cases}
			\frac{1}{2^{q}}\mathbf{U}\mathbf{W}_q\mathbf{S_WT}\mathbf{W}_q\mathbf{P}\mathbf{X},  \ \ \ m \le n\\
			\frac{1}{\sqrt{2^{p+q}}}\mathbf{U}\mathbf{W}_q\mathbf{A_{vg}}\mathbf{S_WT}\mathbf{W}_p\mathbf{P}\mathbf{X}, \ \  m > n
		\end{cases}
	\end{equation}
	In this way, though we double the number of trainable parameters in the Walsh-Hadamard layer, our new layer still contain much fewer trainable parameters than the $1\times1$ convolution layer. We are trying the experiments about the Weighted Smooth-Thresholding currently.

	\section{Experimental Results}
	In our experiments, training and accuracy tests proceed on a laptop with Intel Core i7-7700HQ CPU, NVIDIA GTX-1060 GPU with Max-Q design, and 16GB DDR4 RAM. Code is written in TensorFlow-Keras in Python 3. In the following sections, we will compare the accuracy of neural networks on the Fashion MNIST dataset and CIFAR-10 dataset, then compare the speed of neural networks by feeding a large tensor to a single layer.
	
\begin{figure}[htbp]
		\begin{center}
			\subfloat[]{\includegraphics[width=0.2\linewidth]{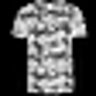}}
			\subfloat[]{\includegraphics[width=0.2\linewidth]{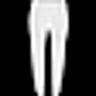}}
			\subfloat[]{\includegraphics[width=0.2\linewidth]{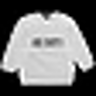}}
			\subfloat[]{\includegraphics[width=0.2\linewidth]{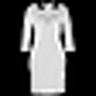}}
			\subfloat[]{\includegraphics[width=0.2\linewidth]{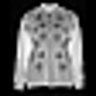}}\\
			\subfloat[]{\includegraphics[width=0.2\linewidth]{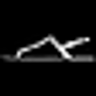}}
			\subfloat[]{\includegraphics[width=0.2\linewidth]{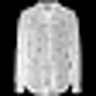}}
			\subfloat[]{\includegraphics[width=0.2\linewidth]{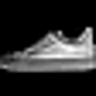}}
			\subfloat[]{\includegraphics[width=0.2\linewidth]{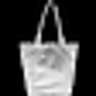}}
			\subfloat[]{\includegraphics[width=0.2\linewidth]{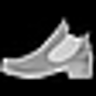}}
		\end{center}
		\caption{Samples of Fashion MNIST: (a) T-shirt/top, (b) trouser, (c) pullover, (d) dress, (e) coat, (f) sandal, (g) shirt, (h) sneaker, (i) bag, (j) ankle boot.}
		\label{fig: fashion MNIST}
	\end{figure}

\begin{figure}[htbp]
		\begin{center}
			\subfloat[]{\includegraphics[width=0.2\linewidth]{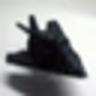}}
			\subfloat[]{\includegraphics[width=0.2\linewidth]{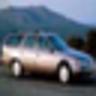}}
			\subfloat[]{\includegraphics[width=0.2\linewidth]{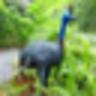}}
			\subfloat[]{\includegraphics[width=0.2\linewidth]{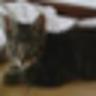}}
			\subfloat[]{\includegraphics[width=0.2\linewidth]{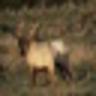}}\\
			\subfloat[]{\includegraphics[width=0.2\linewidth]{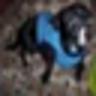}}
			\subfloat[]{\includegraphics[width=0.2\linewidth]{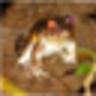}}
			\subfloat[]{\includegraphics[width=0.2\linewidth]{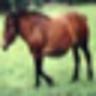}}
			\subfloat[]{\includegraphics[width=0.2\linewidth]{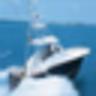}}
			\subfloat[]{\includegraphics[width=0.2\linewidth]{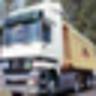}}
		\end{center}
		\caption{Samples of CIFAR-10: (a) airplane, (b) automobile, (c) bird, (d) cat, (e) deer, (f) dog, (g) frog, (h) horse, (i) ship, (j) truck.}
		\label{fig: Cifar-10}
	\end{figure}

	In the accuracy test, we use the ImageNet-pretrained MobileNet-V2 model with 1.0 depth. As it is shown in Table~\ref{tab: model}, We replace layers after the last global average pooling layer by a dropout layer and a dense layer with 10 outputs as TensorFlow official transfer learning demo \cite{Transfer_learning_web}. This is how we build the fine-tuned MobileNet-V2. In case the minimum input of MobileNet-V2 is $96\times96\times3$, we interpolate the images to this resolution, as shown in Figures~\ref{fig: fashion MNIST} and~\ref{fig: Cifar-10}. Because images in Fashion MNIST dataset are in gray-scale, after interpolating, we copy the values 2 times to make them $96\times96\times3$.
	
	\begin{table}[htbp]
		\begin{center}
			\begin{tabular}{|c|c|c|c|c|c|}
				\hline
				Input & Operator & $t$ & $c$ & $n$ & $s$\\
				\hline\hline
				$96^2\times 3$&Conv2D&-&32&1&2\\
				$48^2\times 32$&Bottleneck&1&16&1&1\\
				$48^2\times 16$&Bottleneck&6&24&2&2\\
				$24^2\times 24$&Bottleneck&6&32&3&2\\
				$12^2\times 32$&Bottleneck&6&64&4&2\\
				$6^2\times 64$&Bottleneck&6&96&3&1\\
				$6^2\times 96$&Bottleneck&6&160&3&2\\
				$3^2\times 160$&Bottleneck&6&320&1&1\\
				$3^2\times 320$&Conv2D&-&1280&1&1\\
				$3^2\times 1280$&AvgPool&-&-&1&-\\
				$1280$&Dropout (rate=0.2)&-&-&1&-\\
				$1280$&Dense (units=10)&-&-&1&-\\
				\hline
				\multicolumn{6}{l}{\tabincell{l}{$t$, $c$, $n$ and $s$ represent expansion factor, channel, repeat\\ time and stride \cite{sandler2018mobilenetv2}. Initial weights before Dropout are\\ from ImageNet checkpoint float\_v2\_1.0\_96 in \cite{Mobilenet_web}.}}
			\end{tabular}
		\end{center}	
		\caption{Structure of fine-tuned MobileNet-V2 (baseline)}
		\label{tab: model}
	\end{table}
	
	We modify the MobileNet-V2 with methods described in Section~\ref{Methodology}. Detailly, in Tables~\ref{tab: Fashion MNIST} and~\ref{tab: Cifar-10}, we try to only change $1\times 1$ convolution layers for channel projection with name ``MobileNet-V2 with projection conv. changed" because $1\times1$ convolution layers for channel expansion plays a more important role than $1\times1$ convolution layers for channel projection in bottleneck layers for feature extraction, and we try to change $1\times 1$ convolution layers for both channel expansion and projection with name ``MobileNet-V2 with $1\times 1$ conv. changed", and change $1\times 1$ convolution layers and $3\times 3$ depthwise convolution layers with the name ``MobileNet-V2 with bottleneck changed". Note that early layers contain very few perimeters, but they are much more important in feature extraction than the latter layers, especially for early $1\times1$ convolution layers for channel expansion, so we also try to keep some early layers, as changing these layer cannot save too many parameters but will bring a lot of accuracy loss. For example, if we only change $1\times 1$ convolution layers for both channel expansion and projection in the latter half of bottleneck layers, we call it as ``MobileNet-V2 with 1/2 $1\times 1$ conv. changed".
	
	\begin{table*}[htbp]
		\begin{center}
			\begin{tabular}{|l|c|c|c|c|}
				\hline
				Method&Threshold&Parameters&Saving Rate&Accuracy\\
				\hline\hline
				Fine-tuned MobileNet-V2 (baseline)&-&2,270,794&-&95.37\%\\
				MobileNet-V2 with all projection conv. changed&Soft&1,317,126&42.00\%&94.46\%\\
				MobileNet-V2 with all $1\times 1$  conv. changed&Soft&574,838&74.69\%&91.49\%\\
				MobileNet-V2 with 2/3 $1\times 1$  conv. changed&Soft&616,449&72.85\%&93.90\%\\
				MobileNet-V2 with 1/2 $1\times 1$  conv. changed &Identity$\mathbf{^a}$&716,362&68.45\%&93.45\%\\
				MobileNet-V2 with 1/2 $1\times 1$  conv. changed &ReLU$\mathbf{^b}$&730,648&67.82\%&94.14\%\\
				MobileNet-V2 with 1/2 $1\times 1$  conv. changed&Soft&730,648&67.82\%&94.38\%\\
				\textbf{MobileNet-V2 with 1/2 $1\times 1$  conv. changed} &\textbf{Smooth}&\textbf{730,648}&\textbf{67.82\%}&\textbf{94.49\%}\\
				MobileNet-V2 with 1/2 bottleneck changed&Soft&730,648&67.82\%&94.26\%\\
				\textbf{MobileNet-V2 with 1/2 bottleneck changed}&\textbf{Smooth}&\textbf{730,648}&\textbf{67.82\%}&\textbf{94.44\%}\\
				\hline
				\multicolumn{5}{l}{\tabincell{l}{$\mathbf{^a}$ is achieved without using any function at where smooth-thresholding is in FWHT layers. So there is no\\\ \ trainable variable. Although it reaches 93.45\%, its accuracy plot does not converge like other cases.}}\\
				\multicolumn{5}{l}{$\mathbf{^b}$ is achieved with ReLU instead of smooth-thresholding in FWHT layers. Threshold values are also trainable.}\\
			\end{tabular}
		\end{center}
		\caption{Results on Fashion MNIST Dataset}
		\label{tab: Fashion MNIST}
	\end{table*}
	
	\begin{table*}[htbp]
		\begin{center}
			\begin{tabular}{|l|c|c|c|c|}
				\hline
				Method&Threshold&Parameters&Saving Rate&Accuracy\\
				\hline\hline
				Baseline MobileNet-V2 model$\mathbf{^a}$ in \cite{ayi2020rmnv2}&-&2.2378M&-&94.3\%\\
				RMNv2$\mathbf{^b}$ \cite{ayi2020rmnv2}&-&1.0691M&52.22\%&92.4\%\\
				\hline
				Fine-tuned MobileNet-V2 (baseline)&-&2,270,794&-&95.21\%\\
				MobileNet-V2 with all projection conv. changed&Soft&1,317,126&42.00\%&87.22\%\\
				MobileNet-V2 with 1/2 projection conv. changed&Soft&1,399,328&38.38\%&92.46\%\\
				MobileNet-V2 with 1/3 projection conv. changed&Soft&1,514,036&33.33\%&94.28\%\\
				MobileNet-V2 with 1/2 $1\times 1$ conv.  changed&Soft&730,648&67.82\%&90.16\%\\
				MobileNet-V2 with 1/2 $1\times 1$ conv.  changed&Smooth&730,648&67.82\%&90.91\%\\
				MobileNet-V2 with 1/3 $1\times 1$ conv.  changed&Soft&947,759&58.26\%&93.21\%\\
				\textbf{MobileNet-V2 with 1/3 $1\times 1$ conv.  changed}&\textbf{Smooth}&\textbf{947,759}&\textbf{58.26\%}&\textbf{93.23\%}\\
				MobileNet-V2 with 1/2 bottleneck changed&Soft&730,648&67.82\%&90.11\%\\
				MobileNet-V2 with 1/2 bottleneck changed&Smooth&730,648&67.82\%&90.77\%\\
				MobileNet-V2 with 1/3 bottleneck changed&Soft&947,759&58.26\%&92.47\%\\
				\textbf{MobileNet-V2 with 1/3 bottleneck changed}&\textbf{Smooth}&\textbf{947,759}&\textbf{58.26\%}&\textbf{92.88\%}\\
				\hline
				\multicolumn{5}{l}{Parameters and accuracy of baseline model$\mathbf{^a}$ and RMNv2$\mathbf{^b}$ are from Table 3 in \cite{ayi2020rmnv2}.}
			\end{tabular}
		\end{center}	
		\caption{Results on the CIFAR-10 dataset}
		\label{tab: Cifar-10}
	\end{table*}
	
	\subsection{Fashion MNIST Experiments}
	We train the networks with the same parameter setting (SGD with the learning rate = 0.005, momentum = 0.9 on 100 epochs with batch size = 64) for a fair comparison and record the best accuracy training phase. We take categorical cross-entropy without label smoothing as the loss function. We horizontally flip the training images to enrich the dataset. The results of this experiment are summarized in Table~\ref{tab: Fashion MNIST}. By replacing about a half of $1\times1$ convolution layers with FWHT layers, we can save 67.82\% parameters with 0.98\% accuracy loss. Besides, by also replacing about half of $3\times3$ depthwise separable convolution layers by MF-DS-Conv, we only lose an additional 0.05\% accuracy.

\begin{figure}[htbp]
		\begin{center}
			\includegraphics[width=1\linewidth]{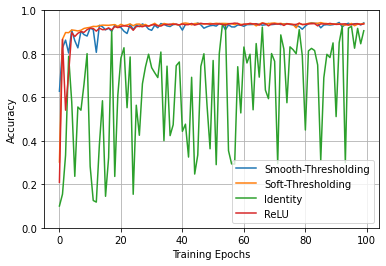}
		\end{center}
		\caption{Accuracy history of networks with different threshold functions in the FWHT layers.}
		\label{fig: acc1}
	\end{figure}

	Moreover, to show the advantage of smooth-thresholding over the soft-thresholding, ReLU ($y=\text{ReLU}(x-T)$) and the linear identity function ($y=x$), we performed a set of experiments in the FWHT layers. Accuracy history is shown in Figure~\ref{fig: acc1}. We only changed the non-linearity in the WH transform domain, and everything else (even the random seed) is identical. If we do not use any function, the network has a very rough converge behavior, as shown in the green curve in Figure ~\ref{fig: acc1}. With ReLU, MobileNet-V2 with the half of $1\times 1$  convolutions changed reaches an accuracy of 0.24\% less than the soft-thresholding-based approach. This is because ReLU is only the non-negative part of the soft-thresholding operator. The positive values are as important as negative values in the WH transform domain. Because of these reasons, soft- and smooth-thresholding can retain more information than ReLU in FWHT layers. On the other hand, with smooth-thresholding, MobileNet-V2 with half of the bottleneck changed reaches a 0.18\% higher accuracy than the soft-thresholding due to the fact that the derivative of the soft-thresholding operator can only be either 1, 0, or -1. 
	
	\subsection{CIFAR-10 Experiments}
	We train the networks with the same parameter setting (SGD with learning rate = 0.005, momentum = 0.9, on 150 epochs with the batch size = 128) for a fair comparison and record the best accuracy in the training phase. We take categorical cross-entropy without label smoothing as the loss function. We horizontally flip, rotate, shift, sheer and zoom the training images to enrich the dataset. By replacing about one-third of $1\times1$ convolution layers with FWHT layers, we can save 58.26\% parameters with 1.98\% accuracy loss. Our network with 947,759 parameters is more accurate (93.23\%) than the comparable slim RMNv2 network (accuracy=92.4\%) described in \cite{ayi2020rmnv2}. 
	
	In addition, by also replacing about one-third of $3\times3$ depthwise separable convolution layers with MF-DS-Conv, we lose an additional 0.35\% accuracy.

	\subsection{The Speed Test}
	We implemented the proposed network on an NVIDIA computer board to show the efficiency of the FWHT layer. As pointed in Section~\ref{FWHT Layer}, the FWHT is an $O(m\log_2 m)$ operation. We compared the $1\times 1$ convolution layer and the FWHT layer by processing a random tensor from $\mathbb{R}^{10\times32\times32\times1024}$ to $\mathbb{R}^{10\times32\times32\times1024}$. 
	We also implemented the Walsh-Hadamard transform (WHT) using ordinary matrix multiplication.
	The only difference between the WHT layer and the FWHT layer is that the former employs matrix-vector-product-based Walsh-Hadamard transform instead of the $O(m\log_2 m)$ fast Walsh-Hadamard transform. The model uploaded to~\cite{Hadamard_speed_code} only contains one layer (FWHT, WHT or $1\times 1$ convolution). The speed tests are performed on the same laptop computer and an NVIDIA Jetson Nano (4GB RAM version). We run the TensorFlow PB model on the laptop, and we convert the PB file to TFLite format for NVIDIA Jetson Nano because it is an ARM-based edge device. The test result is stated in Table~\ref{tab: speed}. There is no doubt that the WHT layer runs the slowest because its complexity is $O(m^2)$. Although the FWHT layer's speed advantage to $1\times1$ convolution layer is not obvious on the laptop computer, the FWHT layer runs about 2 times faster than $1\times1$ convolution layer on the NVIDIA Jetson Nano board. This is because the TFLite model is optimized by TensorFlow, while the PB model is not optimized as much as the TFLite model. 
	
	\begin{table}[htbp]
		\begin{center}
			\begin{tabular}{|l|c|c|}
				\hline
				Layer &On Laptop&On NVIDIA Jetson Nano\\
				\hline\hline
				$1\times 1$ conv.&0.0975s&2.3524s\\
				WHT&0.1064s&5.1580s\\
				\textbf{FWHT}&\textbf{0.0952s}&\textbf{1.1973s}\\
				\hline
			\end{tabular}
		\end{center}
		\caption{Speed test, code is available at~\cite{Hadamard_speed_code}}
		\label{tab: speed}
	\end{table}
	
	\section{Conclusion}
	In this paper, we propose a binary layer called FWHT layer to replace $1\times1$ convolution layer based on the fast Walsh-Hadamard (WH)  transform and smooth-thresholding to change the channel size in deep neural networks. The FWHT layer using an $m\times m$ WH matrix requires only $N$ trainable parameters compared to $1\times1$ convolution layers because the only trainable parameters are the threshold values of the smooth-threshold, which is a tanh-smoothed version of the well-known soft-thresholding operator. The number of parameters $m$ is equal to the channel number or a power of 2 integers larger than the number of channels, and the same threshold values can be used throughout the layer. The WH transform is implemented in $O(m\log_2m)$ arithmetic. As a result, the FWHT layer runs about 2 times faster than the $1\times1$ convolution layer on NVIDIA Jetson Nano with TFLite model format. 
	
	We also establish the relationship between the MF operator and the 2-by-2 WH transform and proposed a depthwise separable multiplication-free convolution layer, a combination of two binary layers.	By using the architecture of MobileNet-V2 with FWHT and MF layers, we reduce the number of parameters of MobileNet-V2 by 67.82\%  with 0.93\% accuracy loss on the Fashion MNIST dataset and 58.26\% savings with 2.33\% accuracy loss on the CIFAR-10 dataset. Although we perform experiments using MobileNet-V2 to compare our results with \cite{ayi2020rmnv2}, similar results can be obtained using other deep neural networks.

	{\small
		\bibliographystyle{unsrt}
		\bibliography{egbib}
	}
	
\end{document}